\documentclass[letterpaper, 10 pt, conference]{ieeeconf}  

\IEEEoverridecommandlockouts                              

\usepackage{graphics} 
\usepackage{epsfig} 
\usepackage{mathptmx} 
\usepackage{times} 
\usepackage{amsmath} 
\usepackage{amssymb}  
\usepackage{graphicx}
\usepackage{caption}
\usepackage{wrapfig}
\usepackage{paralist}
\usepackage{overpic}

\title{\LARGE \bf
MetaView: Few-shot Active Object Recognition}

\author{Wei Wei$^{1}$, Haonan Yu$^{2}$, Haichao Zhang$^{2}$, Wei Xu$^{2}$ and Ying Wu$^{1}$
\thanks{$^{1}$Wei Wei and Ying Wu are with Department of Electrical and Computer Engineering, Northwestern University, Evanston, IL, USA}
\thanks{$^{2}$Haonan Yu, Haichao Zhang and Wei Xu are with Horizon Robotics Inc., Cupertino, CA, USA}
}

\begin{document}

\maketitle
\thispagestyle{empty}
\pagestyle{empty}

\begin{abstract}
In robot sensing scenarios, instead of passively utilizing human captured views, an agent should be able to \textit{actively} choose informative viewpoints of a 3D object as discriminative evidence to boost the recognition accuracy. This task is referred to as \textit{active object recognition}. Recent works on this task rely on a massive amount of training examples to train an optimal view selection policy. But in realistic robot sensing scenarios, the large-scale training data may not exist and whether the intelligent view selection policy can be still learned from few object samples remains unclear. In this paper, we study this new problem which is extremely challenging but very meaningful in robot sensing --- \textit{Few-shot Active Object Recognition}, i.e., to learn view selection policies from few object samples, which has not been considered and addressed before. We solve the proposed problem by adopting the framework of meta learning and name our method ``\textit{MetaView}''. Extensive experiments on both category-level and instance-level classification tasks demonstrate that the proposed method can efficiently resolve issues that are hard for state-of-the-art active object recognition methods to handle, and outperform several baselines by large margins.
\end{abstract}


\section{Introduction}

Visual recognition is an important task of robot learning. For 3D object recognition, the input space can be either in its native 3D format \cite{wu20153d}, like RGB-D and point cloud, or multiple rendered views on 2D images \cite{Su_2015_ICCV}. In robot vision, an agent can be allowed to explore several viewpoints of an object, by rotating the object or camera, for effective recognition in a 3D environment. This is known as the problem of Active Object Recognition (\textbf{AOR}). Different from traditional 3D object recognition \cite{wu20153d}, the input space of AOR is actively acquired and dynamically selected 2D images, making it a more human-like perception task. It is critical to require an agent to actively select informative viewpoints in a limited time budget, to guide better recognition. 

Recently, reinforcement learning (RL) has been applied to AOR problems and achieved promising results \cite{jayaraman2016look,jayaraman2018learning,ramakrishnan-eccv2018,ramakrishnan2019emergence}. Specifically, it learns a policy to select camera movements based on the viewing history. By training such policy, the agent masters the capability of effectively exploring 3D objects. Meanwhile, recurrent neural networks (RNNs) \cite{mnih2014recurrent} are utilized to aggregate the visual information of selected views as an approximation to the object 3D characteristics, and the aggregated information is used for classification. This framework has been adopted as a standard for AOR problems.

Nevertheless, the state-of-the-art AOR methods still have some limitations. First and foremost, they require a massive amount of training samples to train the policy as well as the classifier. On one hand, such large-scale training data with annotations are cumbersome to collect. On the other hand, training samples of classes are not uniformly distributed, and it is common that some classes contain only a few samples \textit{per se} \cite{wang2017learning}. Without large-scale training data, the generalization capabilities of these active recognition systems 
are unclear. 

Another issue of the existing AOR methods is that the testing classes have to be seen during training. In real-world settings, when an agent trained in such a setting encounters new object classes in testing, the recognition system could fail to adapt quickly to recognize them. A naive approach by fine-tuning the pre-trained model on unseen testing classes could also be less reliable if the new classes contain only a few samples (Section \ref{sec5.2}). Therefore, an active recognition system, which can adapt to unseen classes rapidly, is crucial in realistic robot sensing applications.

Furthermore, for instance-level recognition, e.g., when the task is to identify different chairs, can the agent identify the same chair from arbitrary viewpoints, if it is allowed to view that chair only a few glimpses in training? This task is clearly out of the scope of the previous RL-based AOR approaches, since in this case there is only one training sample for each label.

To alleviate the aforementioned issues, we first present a new challenging problem---\textbf{Few-shot Active Object Recognition}, which has not been considered or addressed previously in the AOR literature, but is extremely critical and practical for robotic recognition platforms. Different from the setting of traditional AOR, our AOR problem in the context of the few-shot setting\footnote{Note that in this paper \emph{few-shot} refers to the number of objects for each class in training, rather than the number of views allowed to explore for each object.} does not allow massive training object samples to train the view selection policy and the classifier. In this sense, this new problem cannot be directly resolved by traditional AOR methods. 

Moreover, our new problem focuses on how to train the agent to learn the adaptation of a view selection policy to a new class not seen in training, by using just a handful of training samples. This has also not been explored by traditional AOR methods. In other words, this new problem is targeted on teaching the agent \emph{learning to learn}, \emph{i.e.}, learning a way to learn from few-shot samples of new classes, rather than just \emph{learning to recognize}, which is learning to recognize new objects of several fixed classes, as traditional AOR problem does.

To address the introduced new problem, we propose a new method that trains an AOR system with the few-shot learning and fast adaptation ability, adopting the framework of meta learning. Meta learning is a framework that is designed to learn new concepts rapidly, with only a few new training samples \cite{baxter1998theoretical,baxter2000model,vinyals2016matching,finn2017model}. The idea is that, by training on plenty of recognition tasks in a designed few-shot setting where each task only contains very limited training data, the model is forced to learn an inductive bias, in the form of meta knowledge in the parameter space, for quickly learning new concepts in new tasks.

Specifically, we name the proposed method {\bf MetaView}, which can be viewed as a unified framework that applies meta learning to handle the specific challenge of AOR in few-shot setting. Our method takes a step forward on the basis of the state-of-the-art AOR methods, which are RL-based \cite{jayaraman2016look,jayaraman2018learning,jayaraman2018end}, to validate the possibility of learning an active view selection policy by a few samples. Our method also takes a step forward on the basis of meta learning method, to make it embrace broader and more realistic applications where robot can actively acquire viewpoint with intelligence.
Our method can also address the challenge of fast adaptation to newly observed classes, which is a crucial requirement for robot sensing, while unexplored by AOR methods. 

Overall our major novelty is the introduction of an underexplored problem that is a new appearance of an old problem AOR, in the few-shot setting to fulfill realistic requirements in robot vision scenarios. We believe that this will inspire future works and help push research in this community to a more practical direction. The proposed  method is not a main novelty by itself but is a reasonable initial trial for the introduced problem and may enlighten further research works. To summarize, the main contributions of this work are:

\begin{compactenum}
    \item We present a new problem of AOR in a few-shot learning setting. Such a problem could potentially embrace more realistic applications than the traditional many-shot learning problem. To our best knowledge, this problem has not been investigated before.
    
    \item We explore a solution to the presented problem. We verify that the view selection policy learned from few-shot can boost the recognition accuracy and our trained model has the capability of fast adaptation to new classes. Again to our best knowledge, no existing AOR methods have aimed at this few-shot learning or fast adaptation ability yet.

    \item We adapt existing synthetic and realisitc dataset to evaluate the few-shot learning and fast adaptation ability of the object view selection policy with category-level and instance-level recognition tasks. The modified datasets will be made public for future research on this topic.
\end{compactenum}

\section{Related Work}
\label{related work}

\subsection{Active Object Recognition}


Active object recognition was first introduced in \cite{wilkes1992active}, which developed a system integrating a camera-mounted robot arm with a mobile base. Later, many efforts \cite{dickinson1994active,schiele1998transinformation,borotschnig1998active,pito1999solution,callari2001active,denzler2002information} were spent on optimally selecting viewpoints to reduce ambiguities in visual recognition, by alleviating the issues of occlusion, limited resolution, unfavorable illumination, and other types of degradation. But these methods relied on handcrafted features to solve recognition on small datasets and had difficulties scaling.

RL has been applied on active robot vision and proved effective. Early work can be found in \cite{paletta2000active}, which proposed a system to fuse information by probabilistically encoding 2D views and reinforcing actions that lead to discriminative viewpoints. Embracing the strong representation power of deep networks, \cite{malmir2015deep} applied deep Q-learning to action selection. Nevertheless, the visual feature of each view was pre-trained offline, instead of directly being adjusted along with other modules in an end-to-end manner. In contrast,  \cite{jayaraman2016look,jayaraman2018end} proposed a unified pipeline to simultaneously learn policies for camera movement, single view recognition, and the fusion of multiple selected views, which achieved the state-of-the-art performance in AOR.

Note that all of the previous learning-based AOR methods require large-scale training samples, and could not recognize new object classes without finetuning on many samples of the new classes. This raises an unnegligible practical issue when considering realistic scenarios such as continual online learning and few-shot learning, which motivates our work.

\subsection{Few-shot Learning via Meta learning}
\label{sec:meta_learning}

\begin{figure*}[t]
\vspace{-0mm}
\centering
    \includegraphics[width=\linewidth]{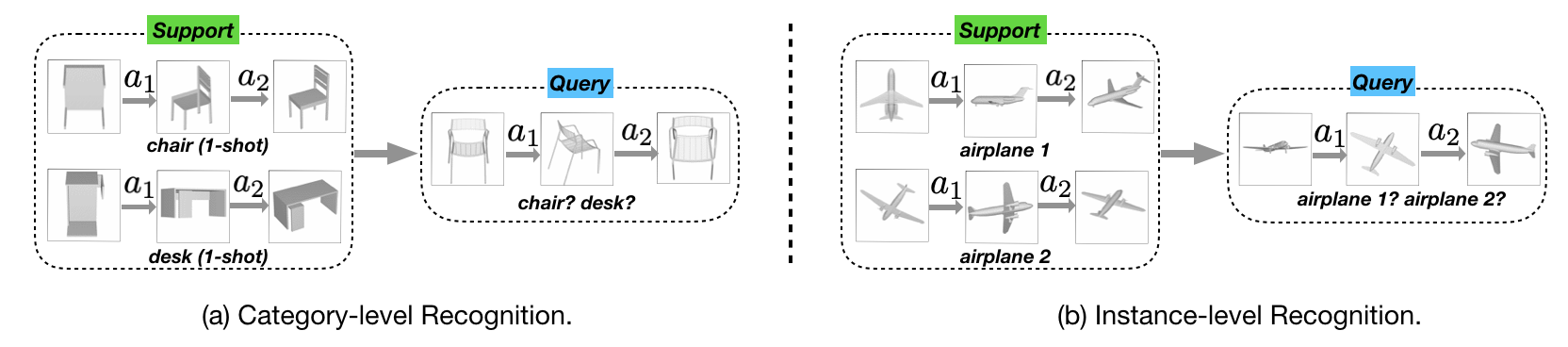}
    \vspace{-0.25in}
\caption{Illustration of the few-shot learning tasks for (a) category-level and (b) instance-level AOR. During the \emph{\scriptsize{\textsf{\textbf{Support}}}} phase, the agent is allowed to view each provided object sample for a limited budget (glimpses) to learn new concepts (category or instance identity) that were never experienced before.  In the \emph{\scriptsize{\textsf{\textbf{Query}}}} stage, given a new instance belonging to one of the concepts encountered in the \emph{\scriptsize{\textsf{\textbf{Support}}}} phase, 
the agent is asked to predict its label given the same budget.}
\label{fig:task}
\vspace{-2mm}
\end{figure*}

Few-shot learning aims to learn and generalize from limited data \cite{fei2006one}. Overfitting is a notorious challenge. Effectively transferring from existing knowledge to learning new few shots is the key. 
Meta learning serves as a good solution. A typical learning machine selects an inductive bias that contains the solution of the task at hand and generalizes well from the task training samples, while a meta learning machine automatically learns an inductive bias from an environment which is assumed to be a task distribution \cite{baxter1998theoretical}. Instead of using a large number of samples to combat overfitting, meta learning samples a large number of tasks, each of which is to learn from few samples. 

Among meta learning methods, MAML \cite{finn2017model} is a popular one which relies on a probabilistic understanding that prior knowledge can be extracted from previous tasks and new tasks can be learned efficiently by Bayesian inference of the posterior. Our method goes further on this basis that the parameter space of Bayesian inference includes not only image representations but also view selection policy parameters. 

This paper emphasizes on learning a view selection policy via meta learning for few-shot 3D object recognition, different from existing meta learning methods which mostly focus on image classification tasks. In Section~\ref{sec:task} we detailedly describe our novel problem and point out two specific task settings. In Section~\ref{sec:meta_view} we show our solution---MetaView, to the proposed problem. In Section \ref{sec:experiments} we report quantitative experiment results in different settings to demonstrate the effectiveness of our method. We conclude and discuss future work in Section \ref{sec:conclusion}.

\section{Tasks}
\label{sec:task}

We present the setup for the underexplored few-shot AOR problem in this section. The task setup is inspired by some common and crucial scenarios that are usually encountered by a 3D robot agent equipped with vision and control systems. The schematic illustration has been provided in Figure~\ref{fig:task}. In general, for one task, during the \emph{support} phase, the agent is allowed to view each provided object sample for a limited number of glimpses to learn the concepts (category or instance identity) that have never been learned before.  Then in the subsequent \emph{query} phase, 
the agent needs to predict the identity of a newly given sample, which belongs to one of the concepts experienced in the support phase, within a number of glimpses. To accomplish these tasks, it is essential for the agent to develop an effective strategy of viewing the object instance in an informative way in both phases given the budget constraint, \emph{i.e., to learn to view 3D objects from few samples}, as addressed by Section~\ref{sec:meta_view}. We further illustrate the task setup with two concrete examples in the following.

The first task is few-shot \emph{category}-level recognition. As illustrated in Figure \ref{fig:task}.(a), suppose that the agent is asked to learn two categories \textit{chair} and \textit{desk} which have never been encountered before. Only a few (\textit{e.g.} 1) instance of either category is allowed to be viewed for a limited  budget (\emph{e.g.} 3 glimpses) in the support stage, within which it's basically impossible to cover and memorize every viewpoint of the instance. In query phase, when a new instance of one of the just learned classes is given (\emph{e.g.}  a new chair instance), can the agent recognize whether it's a chair or a desk, in the same budget by intelligently selecting viewpoints?

The other task is few-shot \emph{instance}-level recognition, as shown in Figure \ref{fig:task} (b). Here the recognition is performed on the instance level in a similar setting as the first task. One difference is that the query instance should be one of the support instances for learning, but starting from arbitrary viewpoints. However, the support instances might be quite similar to each other in appearances, since they come from the same object category. 

We  will present our MetaView approach to solve both tasks in a unified way in next section. 

\section{Approach}
\label{sec:meta_view}

\begin{figure*}[t]
\vspace{-2mm}
\begin{center}
   \includegraphics[width=1\linewidth]{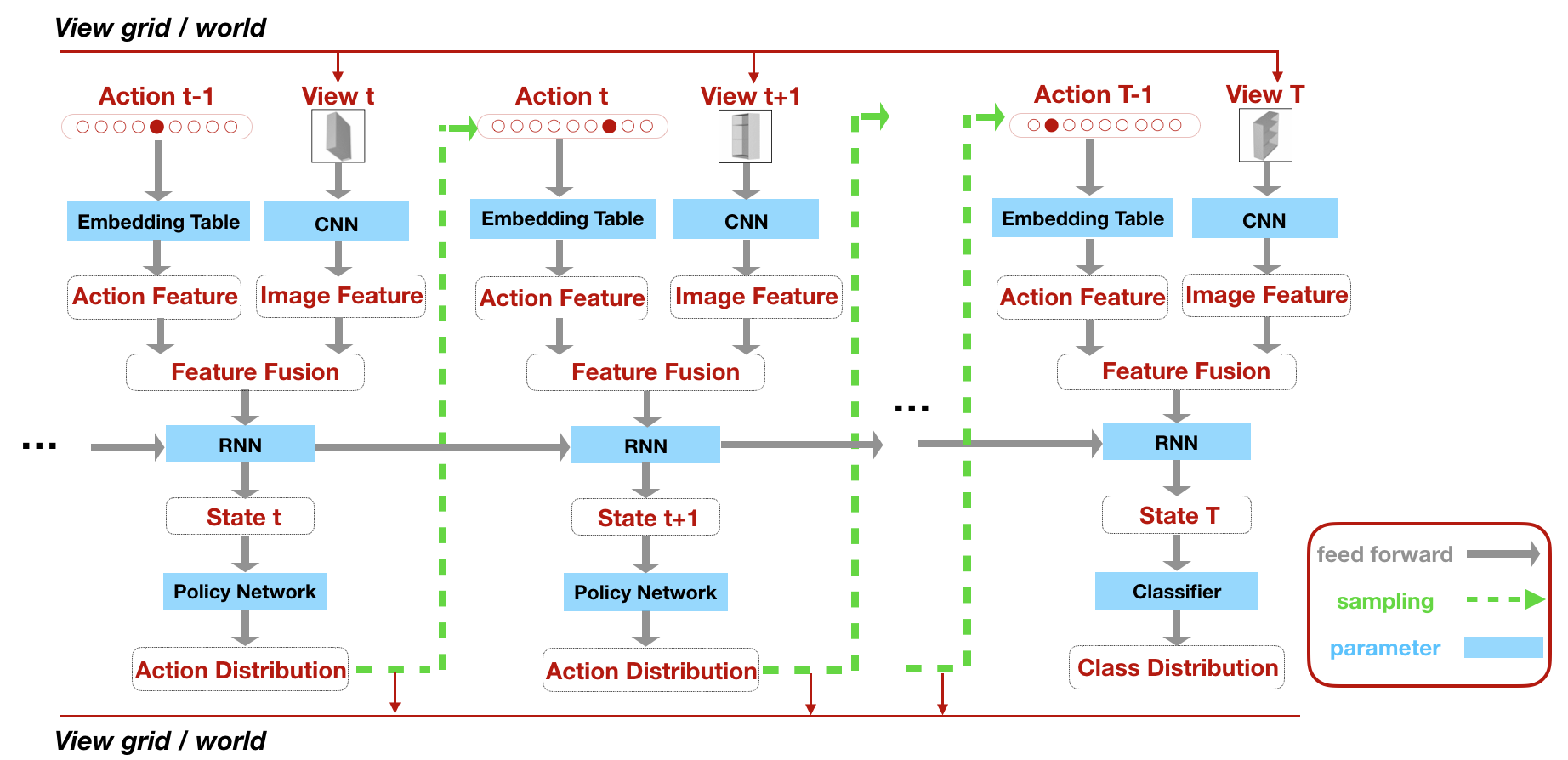}
\end{center}
\vspace{-4mm}
\caption{The pipeline of the active recognition system for a 3D object $x$. A budget of $T$ glimpses are allowed. For each time step, the view image and sampled action (a one-hot vector) constitutes the input observation. Before the last time step, the output is an action distribution, from which the next action is sampled. For the last time step, the output is the predictive class label.}
\label{fig:pipeline}
\vspace{-2mm}
\end{figure*}

Let $x \in X$ denote a 3D object, with $y \in Y$ as its corresponding label (category label or instance label). Different from passive 2D image recognition, a 3D object can be observed from multiple viewpoints that might be selected by the agent. Similar to \cite{jayaraman2016look,jayaraman2018learning,ramakrishnan2019emergence,jayaraman2018end}, we parametrize the viewpoint space with \textit{(elevation, azimuth)} using the polar coordinate system, and for each sampled viewpoint we can render a 2D view image. A view grid which consists of a set of such rendered view images is used to represent the corresponding 3D object. This discretizing manner can be viewed as a reasonable approximation without losing its generality, as also adopted in aforementioned AOR methods. 

The AOR system, as illustrated in Figure~\ref{fig:pipeline}, can be implemented as a parametric mapping from $X$ to $Y$. We require the system to observe each object for only a limited number of time steps ($T$ glimpses). At the final time step $T$, the system outputs a probability distribution of the predicted label for the input object $x$. Before final time step $T$, it can issue an \emph{action} to change the current view to a neighboring one for view exploration. At each time step, the input \emph{observation} contains the rendered 2D view image, as well as the previous action which is defined as the viewpoint shift (\emph{i.e.}, \textit{($\Delta$elevation, $\Delta$azimuth)} ) compared to the previous time step\footnote{For the initial observation, the view is randomly chosen from the view grid and the previous action is simply (0,0).}. The view image is fed into a convolutional neural network (CNN) which is composed of four modules of Conv-BatchNorm-ReLU, to extract its visual features. Meanwhile, the action, which is discrete categorical data,
is converted to a one-hot vector with the dimension same as the total number of available actions, after which a linear layer is applied to encode it into a continuous vector. The reason of using such an action signal as an input at each time step is that it provides a decision context across time steps, potentially help the model to avoid choosing repetitive views in this process. Also, encoding actions as word embeddings \cite{mikolov2013distributed} proves more effective than directly casting them to continuous vectors such as done by previous AOR methods \cite{jayaraman2016look,jayaraman2018end}.

Then the image feature and action embedding are fused into a single vector. To aggregate the information of the current and past observations, an RNN is used to recurrently integrate the current fusion vector with the previous RNN state. The updated RNN state encodes all the past action/view history. For every time step $t<T$, the RNN state which encodes the combined visual features of input views and action embeddings up to $t$, is used to generate an action distribution, from which an action is sampled to change the viewpoint to get the next rendered view image from the environment. For the last time step $T$, the final RNN state, which contains the 3D object representation, is fed to a linear layer to output softmax logits for label prediction.

Every such a feed forward process yields a trajectory denoted by 
$(s_1,a_1,s_2,\cdots,a_{T-1},s_T),$
where $s_t$ and $a_t$ are the RNN state and the view selection action at time $t$, respectively. The loss of a feed forward on an input object consists of three items defined for the model parameters $\theta$ (for notation simplicity, we will ignore $\theta$ in the losses):

1) \emph{Classification loss}, which is the cross entropy between the output distribution $f$ and categorical label $y$, as the standard supervised learning for fitting labels:
\begin{equation}
    \mathcal{L}_{cls} = -\sum_{c=1}^C y_c\log(f(s_T)_c),
    \label{eq:loss_cls}
\end{equation}
where the classifier $f(\cdot)$ outputs one of $C$ classes, and $y_c$ is the $c$-th entry of the one-hot vector $y$.

2) \emph{Policy loss}. Each action $a_t$ is sampled from a distribution based on the state $s_t$. We use a simple policy gradient algorithm REINFORCE \cite{williams1992simple} by updating the parameters to favor the view selection policies that generate higher rewards, by increasing the probability of an action which contributes to a high reward at the final step. The reward $R$ is defined as 1 if the recognition result at the final time step is correct and 0 otherwise. In our case, the REINFORCE loss is simply the sum of log action probabilities weighted by negative rewards, averaged over time steps of a feed forward process:
\begin{equation}
    \mathcal{L}_{policy} = -\frac{1}{T-1}\sum_{t=1}^{T-1} \log\pi(a_t|s_t)R.
    \label{eq:loss_policy} 
\end{equation}
\noindent
3) \emph{Entropy loss}. To encourage exploration and avoid converging to suboptimal policies too early, we want to maximize the entropy of action distribution \cite{mnih2016asynchronous}. This is equivalent to adding a loss to minimize the negative policy entropy:
\begin{equation}
    \mathcal{L}_{ent} = \frac{1}{T-1}\sum_{t=1}^{T-1} \pi(a_t|s_t)\log\pi(a_t|s_t).
    \label{eq:loss_ent} 
\end{equation}
Combining these three losses, the total loss for optimization is as follows:
\begin{equation}
    \mathcal{L} = \mathcal{L}_{cls} + \lambda_1\mathcal{L}_{policy} + \lambda_2\mathcal{L}_{ent}.
    \label{eq:loss}
\end{equation}

After defining the loss of feed forward process on a single object, meta learning \cite{finn2017model} can be utilized to handle few-shot challenge of AOR. We first define a task distribution $P(\mathcal{T})$ that can be used to randomly sample tasks $\mathcal{T}_i$. For each task $\mathcal{T}_i$, there exists an unknown joint distribution $P_{\mathcal{T}_i}(X_i,Y_i)$ and \textit{i.i.d.} data-label pairs $\{(x_j,y_j)\}$ can be sampled from space $(X_i,Y_i)$. Each task $\mathcal{T}_i$ consists of a support set $\mathcal{S}_i = \{(x_j^s,y_j^s)\}$ and a disjoint query set $\mathcal{Q}_i = \{(x_j^q,y_j^q)\}$\footnote{We borrow the terms \emph{support} and \emph{query} from \cite{vinyals2016matching}. The support set of task $\mathcal{T}_i$ is used to learn $\mathcal{T}_i$, and the corresponding query set of $\mathcal{T}_i$ is used to evaluate the performance on this task after acquired the task-level knowledge from its support set.}. To address the few-shot learning problem, the support set of each task has only few instances $x_j^s$ for each given label $y_j^s$. 


More specifically, the meta learning process can be divided into \textsc{MetaTrain}, \textsc{MetaValidation} and \textsc{MetaTest} phases. During \textsc{MetaTrain}, each training task $\mathcal{T}_i^{tr}=(\mathcal{S}_i^{tr},\mathcal{Q}_i^{tr})$ sampled from the task distribution $P(\mathcal{T}|\textsc{MetaTrain)}$ contains a small support set $\mathcal{S}^{tr}_i=\{(x_j^s,y_j^s)\}$, for adapting the parameters from $\theta$ to $\theta_i$ via a single step of gradient descent:
\begin{equation}
   \theta_{i}\gets \theta-\alpha\nabla_{\theta}\mathcal{L}_{\mathcal{S}^{tr}_i}(\theta),
   \label{eq5}
   \vspace{-1mm}
\end{equation}
where $\alpha$ is the learning rate and the loss $\mathcal{L}_{\mathcal{S}^{tr}_i}$ is by averaging the individual object losses defined in Eq. \ref{eq:loss} over the support set $\mathcal{S}^{tr}_i$. The adapted parameters $\theta_i$ are then evaluated on the paired query set $\mathcal{Q}^{tr}_i$ and produce a loss as $\mathcal{L}_{\mathcal{Q}^{tr}_i}(\theta_i)$. After sampling a predefined number of tasks for each epoch, the query set losses are averaged together to actually update $\theta$:
\vspace{-1mm}
\begin{equation}
    \theta\gets\theta-\beta~\nabla_\theta\mathbb{E}_{\mathcal{T}^{tr}_i\sim P(\mathcal{T}|\textsc{MetaTrain})} \Big[\mathcal{L}_{\mathcal{Q}^{tr}_i}(\theta_i)\Big],
    \label{eq6}
\end{equation}
where $\beta$ is the learning rate. After every epoch of \textsc{MetaTrain}, \textsc{MetaValidation} is performed on different tasks $\mathcal{T}_i^{va}=(\mathcal{S}_i^{va},\mathcal{Q}_i^{va})$ sampled from task distribution $P(\mathcal{T}|\textsc{MetaValidation})$ by computing the corresponding query set loss for tuning hyperparameters and preventing overfitting. For evaluation, after \textsc{MetaTrain}, we perform \textsc{MetaTest} on the test tasks $\mathcal{T}_i^{te}=(\mathcal{S}_i^{te},\mathcal{Q}_i^{te})$ sampled from task distribution $P(\mathcal{T}|\textsc{MetaTest})$, and compute the corresponding query set performance. Note that the three phases have no overlapped object labels as required by the few-shot learning setup.

\section{Experiments}
\label{sec:experiments}

In this section, we evaluate the proposed algorithm by \emph{category-level classification} and \emph{instance-level classification} experiments. For the latter, two different settings including \emph{intra-category learning} and \emph{inter-category learning} are used for evaluation.

\subsection{Dataset and Implementation Detail}
\label{sec5.1}

We verify the effectiveness of MetaView using Princeton ModelNet-40 dataset \cite{wu20153d}.
It contains 40 object categories in daily life. The original data is in 3D format for each object. Following the experimental settings in \cite{jayaraman2016look,ramakrishnan2019emergence,jayaraman2018end}, we use 5~elevations $\{0,\pm 30^\circ,\pm 60^\circ\}$ and 6 azimuths $\{0^\circ, 60^\circ, 120^\circ, 180^\circ, 240^\circ, 300^\circ\}$, defined in the polar coordinate system, to render the view grid for each object with code provided by \cite{Su_2015_ICCV}. An overlap between each two neighboring views is preserved to ensure the visual correlation of them. In this way, the agent can make meaningful action decisions. Such simulated environment can 
maximally approximate the perception and manipulation process of a real robot.

We implement our method with PyTorch \cite{paszke2017automatic}. We follow the term \textit{N-way-K-shot} from the meta learning literature \cite{finn2017model}, to denote that each task classifies $N$ classes, each with $K$ samples in the support set. We run 100 epochs in total, with 500 iterations for every epoch. Each iteration is optimized on a mini-batch of two sampled tasks. The learning rates $\alpha$ and $\beta$ are both $10^{-3}$ throughout all experiments.

\subsection{Category-level Classification}
\label{sec5.2}

This experiment aims at classifying different categories with only few training sample. We split the categories of ModelNet-40 into three subsets: 24 categories for \textsc{MetaTrain}, 6 for \textsc{MetaValidation} and 10 for \textsc{MetaTest}. Each sampled task is set to \textit{5-way-1-shot} or \textit{5-way-5-shot}. $\lambda_1$ and $\lambda_2$ are 10 and 0.003 in Eq. \ref{eq:loss}.


For comparison, it is clear that our few-shot AOR problem is a new problem. As it has not been explored yet, existing AOR methods may not solve it \emph{per se}. The closet baseline is the state-of-the-art AOR method called LookAhead \cite{jayaraman2018end}. Since this method is designed to be trained by many shots and tested by data from seen categories in the training, it is difficult to directly apply it to recognizing unseen categories from few shots. We resort to a simple transfer learning strategy. We train a 30-category classifier using all the categories defined in \textsc{MetaTrain} and \textsc{MetaValidation}\footnote{We train by their released code https://github.com/srama2512/visual-exploration/tree/master/src/recognition.}, finetune\footnote{Finetuning was performed by changing the last softmax classifier layer from 1-out-of-30 to 1-out-of-5. Training and finetuning iterations are both 400. The learning rate for finetuning was carefully tuned.} the model on the support set of sampled tasks in \mbox{\textsc{MetaTest}}, and evaluate the finetuned model on the paired query set. 

We also compare with three more baselines. One is to only use the initial randomized view, namely setting $T\!\!=\!\!1$, denoted as RandomOneView, which can be seen as the meta learning baseline. The second baseline uses a non-learnable random view selection policy, denoted as \mbox{RandomMultiView}. The third baseline, denoted as \mbox{LargestMultiView}, chooses the largest allowable action every time. The latter two baselines have the same budget of glimpses as the proposed method, to demonstrate the function of active view selection policy learning.

\begin{table}[!t]
\centering
\caption{Category-level classification accuracy in \textsc{MetaTest}}
\vspace{-2mm}
\label{class-table}
\begin{tabular}{|c | c| c|} 
 \hline
 Method & \textit{5way-1shot} & \textit{5way-5shot}  \\ 
 \hline\hline
 LookAhead \cite{jayaraman2018end} (T=3) & 38.67\%  &  57.33\% \\
 RandomOneView (T=1) &  43.84\%  & 54.44\% \\ 
 RandomMultiView (T=3) & 50.67\% & 70.42\% \\
 LargestMultiView (T=3) & 53.59\% &  68.76\% \\
 MetaView (T=3)  &  \bf{59.77\%}  & \bf{74.54\%} \\
 \hline
\end{tabular}
\vspace{-2mm}
\end{table}

For LookAhead, we found that the accuracy of classifying 30 categories during training can reach 73.2\%. However, finetuning on the support sets of sampled tasks in \textsc{MetaTest} only yield average accuracies of classifying 5 new categories as 38.67\% and 57.33\% (Table \ref{class-table}). We conjecture this may be partly due to the imbalanced dataset sizes in the training and finetuning stages. The model pretrained on a large dataset could hardly be transferred to learning to classify new categories from only few training samples. To simply verify this conjecture, we also train MetaView so that each sampled task in \textsc{MetaTrain} is not one-shot but many-shot (\textit{e.g.}, 20). In this case, if the support set of \textsc{MetaTest} is also 20-shot, the query accuracy for \textit{5way-1shot} is 71.45\%. However, the performance degrades significantly if the support set of \textsc{MetaTest} is one-shot (query accuracy decreasing to 43.59\%). This could imply from a side why simply finetuning the existing AOR method (trained by massive data) on few-shot to solve the proposed few-shot-AOR problem will suffer in performance.

 


From Table \ref{class-table}, compared to all baselines, the action policy can be trained to play a significant role in contributing to the performance improvement.

\subsection{Instance-level Classification}
\label{sec5.3}

\subsubsection{Intra-category Learning}
\label{sec5.3.1}

This experiment aims at few-shot learning for recognizing new instances within a category. In this setting, we choose one specific category from ModelNet-40 with the smallest appearance variation---``airplane". It contains 726 different instances in total, which we split into three disjoint subsets: 400 instances for \textsc{MetaTrain}, 126 for \textsc{MetaValidation}, and 200 for \textsc{MetaTest}. 

\begin{figure}[!t]
\begin{center}
\includegraphics[width=1.\linewidth]{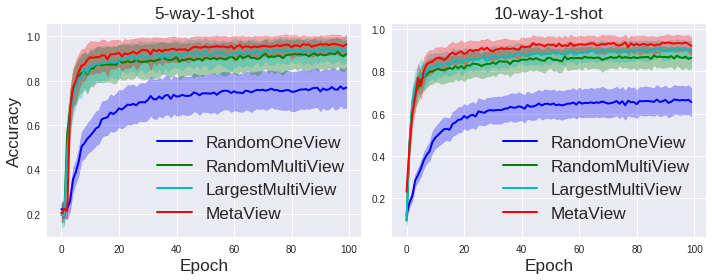}
\end{center}
\vspace{-5mm}
\caption{Accuracy curves of intra-category learning. The \textsc{MetaTrain} and \textsc{MetaValidation} processes are represented by dashed and solid lines respectively. The shading areas show the standard deviations of validation accuracies.}
\label{inner_fig}
\end{figure}

We design two experimental settings: \textit{5-way-1-shot} and \textit{10-way-1-shot}. In each setting, the query set has the same instances as in the support set, but the initial views are all different. The challenge is that instances within a category may be difficult to be distinguished with each other. One can imagine that some subtle differences could possibly \textit{hide} in certain views. 

The agent has a chance of viewing instances in the support set, each instance for only $T$ glimpses. This requires it to extract discriminative features as an approximation to each 3D object representation in a short time. After learning on the support set, when given an instance with a random initial view, the agent should know how to rotate the object or manipulate the viewpoints to match the views seen in the support set, in order to classify the object identity correctly, as illustrated in Figure~\ref{fig:task}.(b). 

\begin{table}[!t]
\centering
\caption{\textsc{MetaTest} accuracy of intra-category learning}
\vspace{-2mm}
\label{inner_table}
\begin{tabular}{|c | c| c|} 
 \hline
 Task & \small{\textit{5-way-1-shot}} & \small{\textit{10-way-1-shot}}  \\ 
 \hline\hline
 \small{RandomOneView} ($T$=1) &  75.24\%  & 65.32\%\\ 
 \small{RandomMultiView} ($T$=3) & 91.32\%  & 85.10\% \\
 \small{LargestMultiView} ($T$=3) & 93.03\%  & 88.03\% \\
 MetaView ($T$=3) &  \bf{94.65\%}  & \bf{91.73\%} \\
 \hline
\end{tabular}
\vspace{-4mm}
\end{table}

Instance-level classification was not studied previously for AOR methods, therefore we compare MetaView with the three baselines in Table \ref{inner_table}. Figure \ref{inner_fig} shows the accuracy curves in \textsc{MetaTrain} and \textsc{MetaValidation}. From both, we see that MetaView boosts the recognition accuracy over two baselines which do not train policies. The performance of meta learning baseline (RandomOneView) is worse than the other three by large margins, indicating that multiple views are necessary for better recognition. For this particular task, $T$ random or largest views already contain the majority of discriminative information for recognition, as demonstrated by the performance of RandomMultiView and LargestMultiView. Even on such high baselines, MetaView still improves the accuracy, showing the effectiveness of active view selection, since the taskg gets much more difficult approaching 100\% (for handling long-tail cases).

\subsubsection{Inter-category Learning}
\label{sec5.3.2}

In this experiment, we remove the restriction of using one specific object category as in the previous intra-category learning experiment. We will still classify instance identities in every task, but \textsc{MetaTrain} and \textsc{MetaTest} contain instances from different object categories. In this setting, we use all 40 categories in ModelNet-40. We choose 24 categories for \textsc{MetaTrain}, 6 categories for \textsc{MetaValidation}, and 10 categories for \textsc{MetaTest}, without any overlapping. For each task, one category is first sampled, after which instances of that category are sampled. The support and query sets contain the same sampled instances, but the initial views are all different. 

\begin{figure}[!t]
\begin{center}
\includegraphics[width=1.\linewidth]{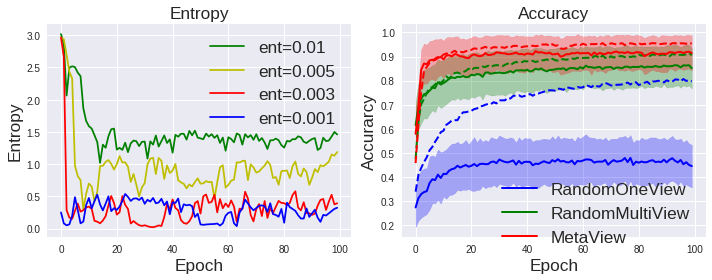}
\end{center}
\vspace{-6mm}
\caption{Training curves for inter-category learning. Left: policy entropy curves with different values of the entropy coefficient $\lambda_2$. Right: training accuracy curves with $\lambda_2\!\!=\!\!0.003$. \textsc{MetaTrain} and \textsc{MetaValidation} are denoted by dashed and solid lines, respectively. The shading areas represent the standard deviations of validation accuracies.}
\label{inter-fig}
\end{figure}

\begin{table}
\centering
\caption{\textsc{MetaTest} accuracy for inter-category learning}
\vspace{-2mm}
\label{inter-table}
\begin{tabular}{|c | c| c| c| c|} 
 \hline
 Task & \textit{5-way-1-shot}   \\ 
 \hline\hline
 RandomOneView ($T$=1) &  60.32\% \\ 
 RandomMultiView ($T$=3) & 84.09\%   \\
 LargestmMultiView ($T$=3) & 86.29\%   \\
 MetaView ($T$=3,$\lambda_2$=0.01) &  92.33\%   \\
 MetaView ($T$=3,$\lambda_2$=0.005) &  90.90\%   \\
 MetaView ($T$=3,$\lambda_2$=0.003) &  \bf{92.78\%}   \\
 MetaView ($T$=3,$\lambda_2$=0.001) &  90.55\%   \\
 \hline
\end{tabular}
\vspace{-4mm}
\end{table}

This task is designed to test if the agent has the capability of learning to recognize object instances of category ``A'' (\textit{e.g.}, ``chair'') from few shots, after having been \emph{only} trained to recognize object instances of category ``B'' (\textit{e.g.}, ``desk'') from few shots.

\begin{figure*}[!t]
\begin{center}
	\begin{overpic}[width=0.76\linewidth]{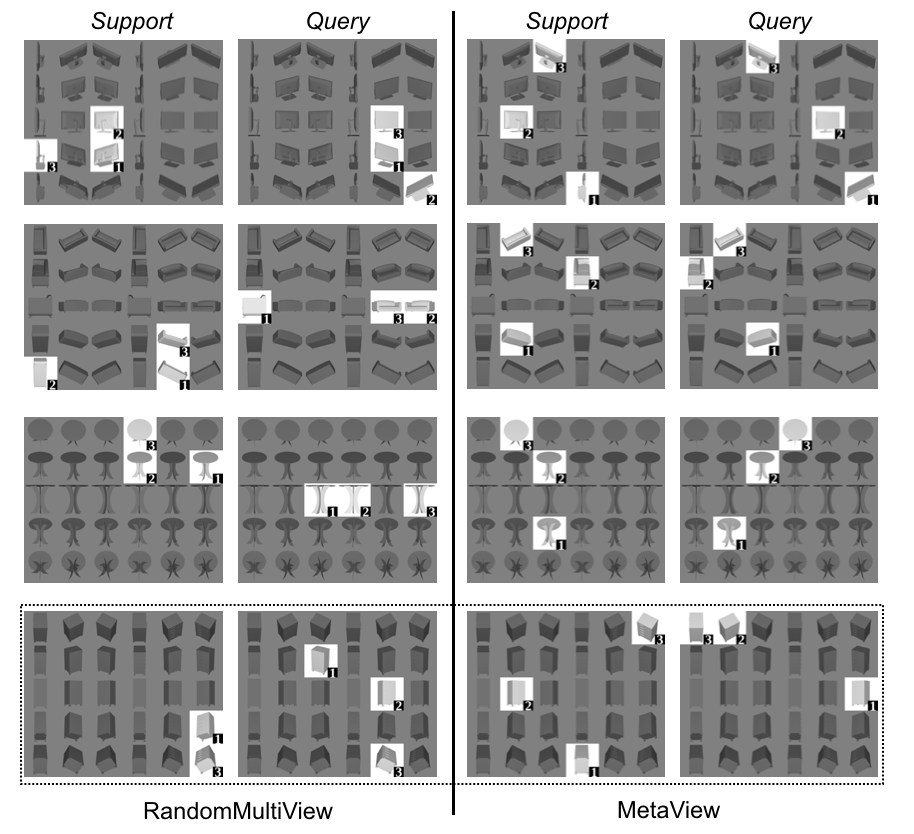}
	\end{overpic}
\end{center}
\vspace{-6mm}
\caption{Visualizations of the view selection trajectories of RandomMultiView and MetaView during \textsc{MetaTest} for inter-category learning. The brighter views represent the selected ones. The time steps are marked at bottom-right corners of selected views. The last row in the dashed rectangle illustrates a case in which our learned policy was not able to select reasonable views for the query instance.} 
\vspace{-0.1in}
\label{fig:visualization}
\end{figure*}

Still this task is out of the scope of traditional AOR methods. We compare MetaView with the three baselines on \textit{5-way-1-shot}. Figure \ref{inter-fig} shows the accuracy curves in \textsc{MetaTrain} and \textsc{MetaValidation}. Table \ref{inter-table} shows the accurices in \textsc{MetaTest}. From both,
we can see that the proposed method is better than the baselines significantly, regardless of the choice for the entropy coefficient $\lambda_2$ in Eq.~\ref{eq:loss}. Note that the performance of the proposed method can be tuned even better with a relatively moderate $\lambda_2$ value of $0.003$ which strikes a good balance between policy exploration and exploitation.

For a comprehensive analysis, we visualize several examples of view selection trajectories of RandomMultiView and MetaView in Figure \ref{fig:visualization}. Generally, for MetaView the selected views in each support set tend to cover more diverse visual appearances of the object, and the selected views in the paired query set tend to (partially) match those in the support set. In contrast, for RandomMultiView there is no guarantee for the diversity and the correlation between views in a query set and those in the support set is much less. We also notice that the learned policy does not always perform reasonably well on every sampled task, as shown in the last row of Figure \ref{fig:visualization}. How to improve upon the current policy training is left for future work.

\subsection{Experiment on Realistic Data}
\label{sec5.4}

\begin{table}[!t]
\vspace{-2mm}
\centering
\caption{Category-level classification accuracy in \textsc{MetaTest} on a realistic dataset SUN360 \cite{xiao2012recognizing}.}
\vspace{-2mm}
\label{real-table}
\begin{tabular}{|c | c| c|} 
 \hline
 Method & \textit{5way-1shot} & \textit{5way-5shot}  \\ 
 \hline\hline
 LookAhead \cite{jayaraman2018end} (T=3) & 24.70\%  &  36.22\% \\
 RandomOneView (T=1) & 28.13\%  & 34.92\% \\ 
 RandomMultiView (T=3) & 33.19\% & 39.15\% \\
 LargestMultiView (T=3) & 34.17\% &  40.05\% \\
 MetaView (T=3)  &  \bf{42.11\%}  & \bf{47.53\%} \\
 \hline
\end{tabular}
\vspace{-4mm}
\end{table}

Although multiple experiments of different settings on ModelNet-40 dataset have demonstrated the effectiveness of MetaView, one would question whether such improvements can be transferred to the real world. To resolve this  doubt, we conduct an experiment with a realistic dataset. Since real data which fits in our proposed task setting is currently unavailable, and COVID-19 prevents us from collecting a realistic dataset in the lab, we test on a similar but slightly different task --- few-shot scene categorization using SUN360 dataset \cite{xiao2012recognizing}. The task is to identify the category of the scene with only few glimpses of local patches of the entire panorama. For each sampled task, few-shot (1 or 5) panoramas are given for training. The viewgrid is larger than ModelNet-40 experiments with 8 azimuths and 4 elevations. The rest of settings and hyperparameters are same with Section \ref{sec5.2}.

In Table \ref{real-table}, again MetaView outperforms the finetuned version of LookAhead \cite{jayaraman2018end} and the meta learning baselines without policy training. This indicates that the proposed method is a promising way to solve real-world robot sensing tasks.

\section{Conclusion}
\label{sec:conclusion}
\label{sec:conclusion}

In this work, we have revisited the classic active object recognition problem and raised a novel problem of AOR in the context of few-shot learning setting, inspired by common but critical scenarios that are usually encountered by robots in practice. We have presented a meta learning approach MetaView to learn to view new 3D objects from few samples, which significantly overcomes the limitations of state-of-the-art AOR methods in our novel problem.
The proposed approach can learn to actively select informative views from the support and query set for better recognition. Extensive experiments on category-level and instance-level classification under different settings compared with a number of baseline approaches have demonstrated the effectiveness of our approach. 

The simulation in our experiments can be seamlessly applied to real robots by setting up a camera sensor and action manipulation program. Also the few-shot settings can efficiently ease the difficulty of data collection in reality. After thorough analysis of few-shot generalization performance in the simulated environment, deploying our method on a robot arm for real cases will be our next move.

Although designed to be closer to the deployment in realistic scenarios ({\emph{e.g.}}, a 3D robot with vision and control) than the passive approaches, the proposed method still faces a number of challenges towards this goal. For example, how to construct a support set from a continuous stream of visual perceptions in real world, and how to learn more concepts continuously.
Recent progresses on external memory~\cite{NTM} and life-long learning~\cite{life_long,Wu_2019_CVPR} could possibly shed lights on these challenges. 
We view this work as one of the earliest attempts in this challenging but promising direction, and hope that this work can contribute new perspectives and encourage more future efforts on this problem.

\addtolength{\textheight}{-12cm}   








\bibliographystyle{IEEEtran.bst}
\bibliography{IEEEexample}  

\end{document}